\documentclass[11pt,a4paper]{article}
\usepackage[hyperref]{naaclhlt2019}
\usepackage{times}
\usepackage{latexsym}
\usepackage{xcolor}
\usepackage{amsmath}

\usepackage[draft]{todonotes}

\usepackage{url}

\aclfinalcopy 

\title{Non-Parametric Adaptation for Neural Machine Translation}

\author{Ankur Bapna \\
  Google AI \\
  {\tt ankurbpn@google.com} \\\And
  Orhan Firat \\
  Google AI \\
  {\tt orhanf@google.com} \\}

\date{}

\begin{document}
\maketitle
\begin{abstract}
Neural Networks trained with gradient descent are known to be susceptible to catastrophic forgetting caused by parameter shift during the training process. In the context of Neural Machine Translation (NMT) this results in poor performance on heterogeneous datasets and on sub-tasks like rare phrase translation. On the other hand, non-parametric approaches are immune to forgetting, perfectly complementing the generalization ability of NMT. However, attempts to combine non-parametric or retrieval based approaches with NMT have only been successful on narrow domains, possibly due to over-reliance on sentence level retrieval. We propose a novel n-gram level retrieval approach that relies on local phrase level similarities, allowing us to retrieve neighbors that are useful for translation even when overall sentence similarity is low. We complement this with an expressive neural network, allowing our model to extract information from the noisy retrieved context. We evaluate our semi-parametric NMT approach on a heterogeneous dataset composed of WMT, IWSLT, JRC-Acquis and OpenSubtitles, and demonstrate gains on all 4 evaluation sets. The semi-parametric nature of our approach opens the door for non-parametric domain adaptation, demonstrating strong inference-time adaptation performance on new domains without the need for any parameter updates.
\end{abstract}

\section{Introduction}
Over the last few years, neural sequence to sequence models \cite{sutskever2014sequence,BahdanauCB15,cho2014learning} have revolutionized the field of machine translation by significantly improving translation quality over their phrase based counterparts \cite{sennrich2015improving,wu2016google,DBLP:journals/corr/ZhouCWLX16}. With more gains arising from continued research on new neural network architectures and accompanying training techniques \cite{DBLP:journals/corr/VaswaniSPUJGKP17,DBLP:journals/corr/GehringAGYD17,chen2018best}, NMT researchers, both in industry and academia, have doubled down on their ability to train high capacity models on large corpora with gradient based optimization.

However, despite huge improvements in overall translation quality NMT has shown some glaring weaknesses, including idiom processing, and rare word or phrase translation \cite{koehn2017six,isabelle2017challenge,lee2019hallucinations} - tasks that should be easy if the model could retain learned information from individual training examples. NMT has also been shown to perform poorly when dealing with multi-domain data \cite{farajian2017neural}. This `catastrophic forgetting' problem has been well-studied in traditional neural network literature, caused by parameter shift during the training process \cite{mccloskey1989catastrophic,santoro2016one}. Non-parametric methods, on the other hand, are resistant to forgetting but are prone to over-fitting due to their reliance on individual training examples. We focus on a non-parametric extension to NMT, hoping to combine the generalization ability of neural networks with the eidetic memory of non-parametric methods. Given a translation query, we rely on an external retrieval mechanism to find similar source-target instances in the training corpus, which are then utilized by the model.

There has been some work on semi-parametric NMT \cite{gu2017search,zhang2018guiding,D18-1340}, but its effectiveness has been confined to narrow domain datasets. Existing approaches have relied on sentence level similarity metrics for retrieval, which works well for domains with high train-test overlap, but fails to retrieve useful candidates for broad domains. Even if we could find training instances with overlapping phrases it's likely that the information in most retrieved source-target pairs is noise for the purpose of translating the current query.

To retrieve useful candidates when sentence similarity is low, we use n-gram retrieval instead of sentence retrieval. This results in neighbors which have high local overlap with the source sentence, even if they are significantly different in terms of overall sentence similarity. This is intuitively similar to utilizing information from a phrase table \cite{koehn2003statistical} within NMT \cite{dahlmann2017neural}, without losing the global context lost when constructing the phrase table. We also propose another simple extension using dense vectors for n-gram retrieval which allows us to exploit similarities beyond lexical overlap.

To effectively extract the signal from the noisy retrieved neighbors, we develop an extension of the approach proposed in \cite{D18-1340}. While \cite{D18-1340} encode the retrieved targets without any context, we incorporate information from the current and retrieved sources while encoding the retrieved target, in order to distinguish useful information from noise.

We evaluate our semi-parametric NMT approach on two tasks.

\begin{itemize}
    \item We evaluate our approach on a multi-domain English-French corpus constructed from narrow domain datasets like JRC-Acquis \cite{steinberger2006jrc,tiedemann2012parallel}  and OpenSubtitles \cite{tiedemann2009news}\footnote{http://www.opensubtitles.org/}, and the standard IWSLT and WMT bilingual corpora, as described in Sections \ref{sec_exp} and \ref{sec_res}. Our results, for the first time, indicate that semi-parametric NMT can be beneficial beyond narrow domain tasks, demonstrating gains of around 0.5 BLEU on WMT, and huge gains ranging from 2-10 BLEU points on IWSLT, JRC-Acquis and OpenSubtitles, when compared to a strong sequence to sequence baseline.
\item The semi-parametric nature of our model enables non-parametric inference-time adaptation to new datasets, without the need for any parameter updates. When trained on WMT and evaluated on the other datasets, our model out-performs fine-tuning based adaptation \cite{luong2015stanford} on JRC-Acquis and OpenSubtitles, and significantly improves performance over the non-adapted model on IWSLT.
\end{itemize}
\section{Semi-parametric NMT}
Standard approaches for Neural Machine Translation rely on seq2seq architectures \cite{sutskever2014sequence,BahdanauCB15}, where given a source sequence $X=\{x_1, x_2, \ldots x_{T_x}\}$ and a target sequence $Y=\{y_1, y_2, \ldots y_{T_y}\}$, the goal is to model the probability distribution, $p(y_t|X, y_1, \ldots y_{t-1})$.

Semi-parametric NMT \cite{dahlmann2017neural,gu2017search} approaches this learning problem with a different formulation, by modeling $p(y_t|X, y_1, \ldots y_{t-1}, \Phi_X)$ instead, where $\Phi_X = \{(X^1, Y^1) \ldots (X^N, Y^N)\}$ is the set of sentence pairs where the source sentence is a neighbor of $X$, retrieved from the training corpus using some similarity metric. This relies on a two step approach - the retrieval stage finds training instances, $(X^i, Y^i)$, similar to the source sentence $X$, and the translation stage generates the target sequence $Y$ given $X$ and $\Phi_X$. We follow this setup, proposing improvements to both stages in order to enhance the applicability of semi-parametric NMT to more general translation tasks.

\subsection{Retrieval Approaches}
Existing approaches have proposed using off the shelf search engines for the retrieval stage. However, our objective differs from traditional information retrieval, since the goal of retrieval in semi-parametric NMT is to find neighbors which might improve translation performance, which might not correlate with maximizing sentence similarity.

Our baseline strategy relies on a sentence level similarity score, similar to those used for standard information retrieval tasks \cite{robertson2004understanding}. We compare this against finer-grained n-gram retrieval using the same similarity metric. We also propose a dense vector based n-gram retrieval strategy, using representations extracted from a pre-trained NMT model.

\subsubsection{IDF Based Sentence Retrieval}
Our baseline approach relies on a simple inverse document frequency (IDF) based similarity score. We define the IDF score of any token, $t$, as $f_t = \log(\frac{\|C\|}{n_t})$, where $\|C\|$ is the number of sentence pairs in training corpus and $n_t$ is the number of sentences $t$ occurs in. Let any two sentence pairs in the corpus be $(X^i, Y^i)$ and $(X^j, Y^j)$. Then we define the similarity between $(X^i, Y^i)$ and $(X^j, Y^j)$ by,
\begin{equation}
sim(X^i, X^j) = 2 \times \Sigma_{t \in (X^i \cap X^j)} f_t - \Sigma_{t \in (X^i \cup X^j)} f_t
\end{equation}
For every sentence in the training, dev and test corpora, we find the $N$ most similar training sentence pairs and provide them as context to NMT.

\subsubsection{IDF Based N-Gram Retrieval}
Motivated by phrase based SMT, we retrieve neighbors which have high local, sub-sentence level overlap with the source sentence. We adapt our approach to retrieve n-grams instead of sentences. We note that the similarity metric defined above for sentences is equally applicable for n-gram retrieval.

Let $X = (t^1, ... t^T)$ be a sentence. Then the set of all possible n-grams of X, for a given $n$, can be defined as $S_X^n = \{(t_i, ... t_{i+n}) \, \forall \, 1 \leq i \leq T \}$ (also including padding at the end). To reduce the number of n-grams used to represent every sentence, we define the reduced set of n-grams for X to be $\hat{S_X^n} = \{(t_i, ... t_{i+n}) \, \forall \, 1 \leq i \leq T, \, i \mod{\frac{n}{2}} = 1 \}$.

We represent every sentence by their reduced n-gram set. For every n-gram in $\hat{S_X^n}$, we find the closest n-gram in the training set using the IDF similarity defined above. For each retrieved n-gram we find the corresponding sentence (In case an n-gram is present in multiple sentences, we choose one randomly). The set of neighbors of $X$ is then the set of all sentences in the training corpus that contain an n-gram that maximizes the n-gram similarity with any n-gram in $\hat{S_X^n}$.

To capture phrases of different lengths we use multiple n-gram widths, $n$. In case a sentence has already been added to the retrieved set, we find the next most similar sentence to avoid having duplicates. The number of neighbors retrieved for each source sentence is proportional to its length.

\subsubsection{Dense Vector Based N-Gram Retrieval}
We also extend our n-gram retrieval strategy with dense vector based n-gram representations. The objective behind using a dense vector based approach is to incorporate information relevant to the translation task in the retrieval stage. We use a pre-trained Transformer Base \cite{DBLP:journals/corr/VaswaniSPUJGKP17} encoder trained on WMT to generate sub-word level dense representations for the sentence. The representation for each n-gram is now defined to be the mean of the representations of all its constituent sub-words. We use the $L2$ distance of n-gram representations as the retrieval criterion. Note that we use a sub-word level decomposition of sentences for dense retrieval, as compared to word-level for IDF based retrieval (i.e., n-grams are composed of sub-words instead of words).

Following the approach described for IDF based n-gram retrieval, we use multiple values of $n$, and remove duplicate neighbors while creating the retrieved set.

\begin{figure}[h]
\includegraphics[scale=0.18]{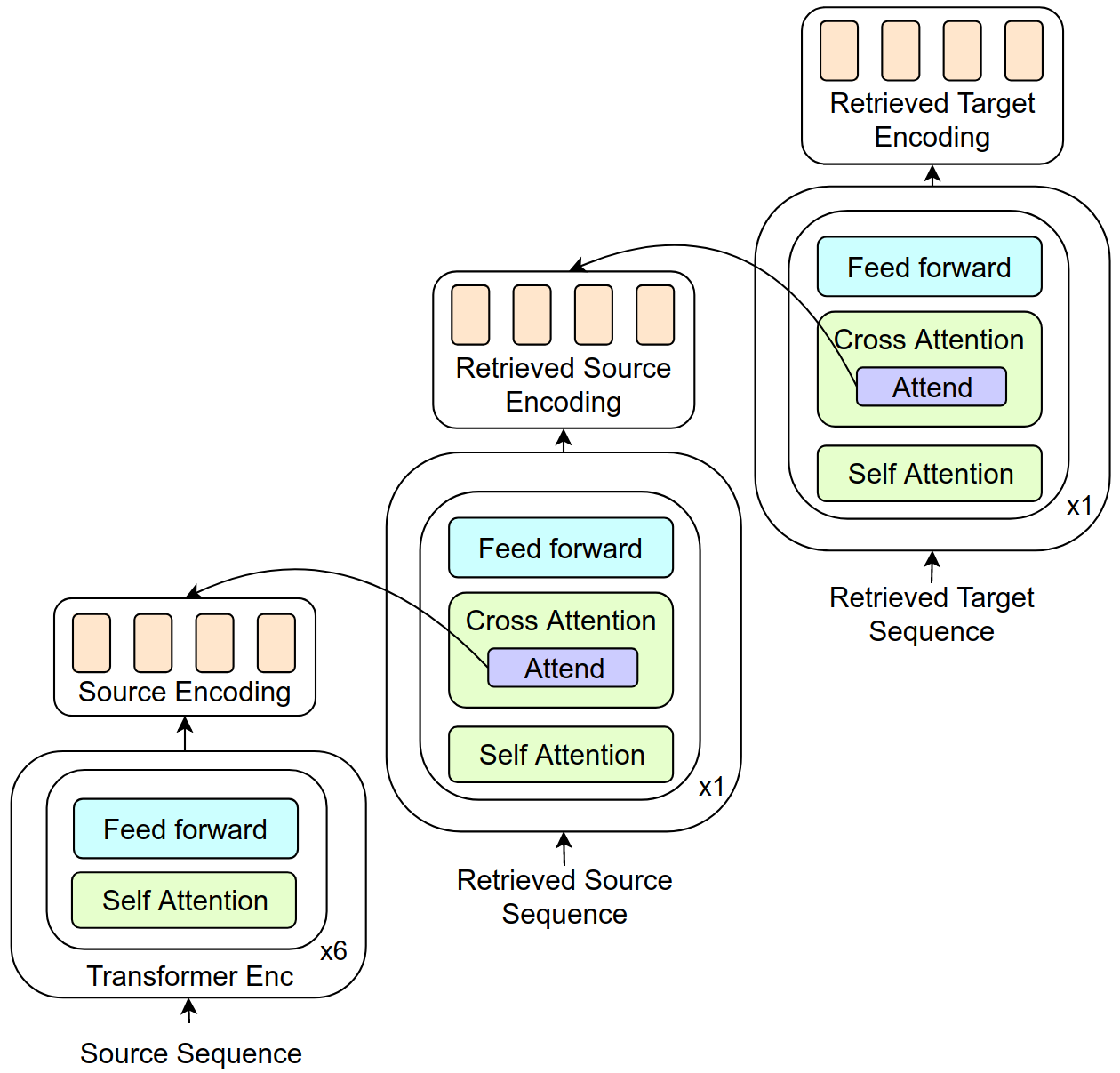}
\caption{Architecture of the Conditional Source Target Memory. The retrieved targets, $Y^i$, are encoded in a transformer encoder, incorporating the attention context from the retrieved sources, $X^i$. In turn, the retrieved sources, $X^i$, are encoded while incorporating context from the current translation source, $X$.}
\label{fig:cstm}
\end{figure}
\begin{figure*}[h]
\centering
\includegraphics[scale=0.25]{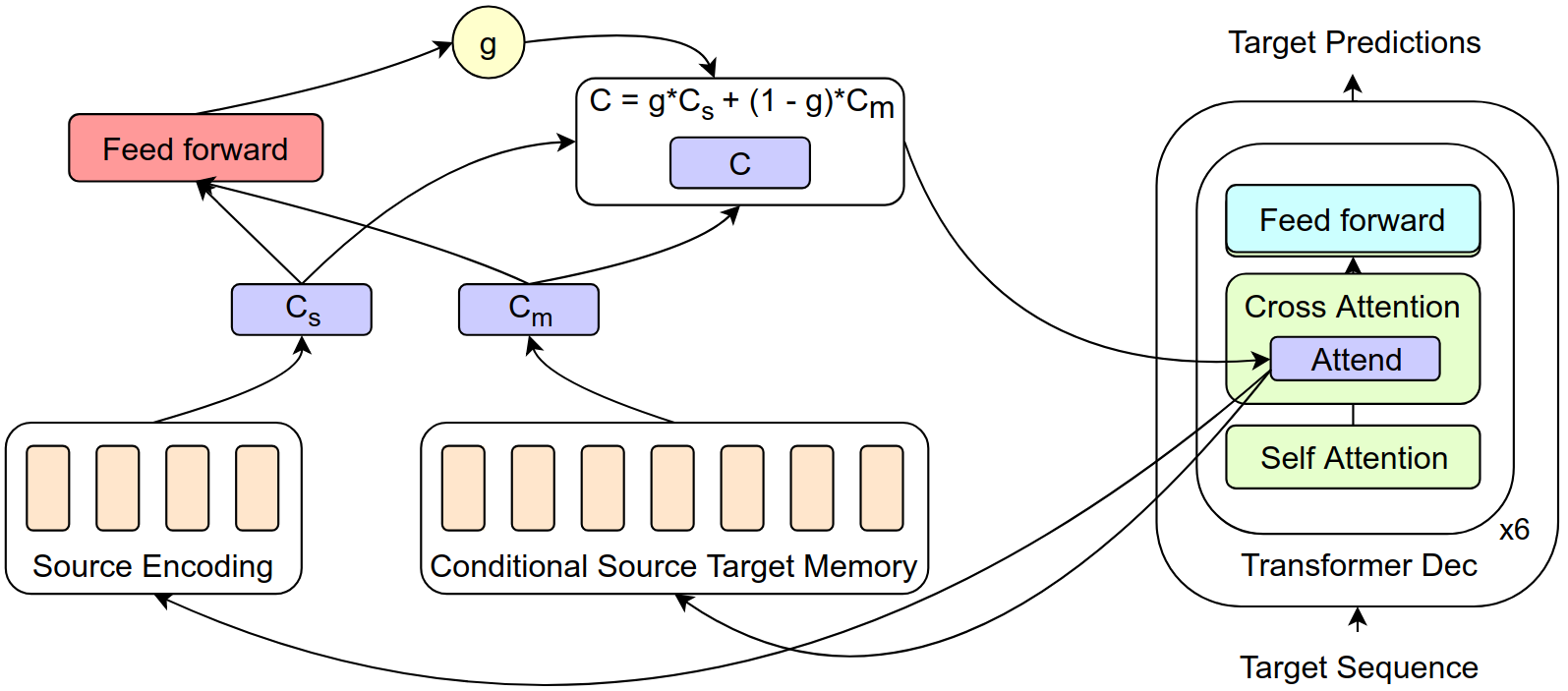}
\caption{Architecture of the gated attention mechanism used in the multi-source transformer decoder.}
\label{fig:gmsd}
\end{figure*}

\subsection{NMT with Context Retrieval}
To incorporate the retrieved neighbors, $\Phi_X$, within the NMT model, we first encode them using Transformer layers, as described in subsection \ref{subsubsec_cstm}. This encoded memory is then used within the decoder via an attention mechanism, as described in subsection \ref{subsubsec_gmsd}.

\subsubsection{Conditional Source Target Memory}
\label{subsubsec_cstm}
We now describe how each retrieved translation pair, $(X^i, Y^i)$, is encoded. This architecture is illustrated in Figure~\ref{fig:cstm}.
\begin{itemize}
\item We first encode the retrieved source, $X^i$, in a Transformer layer. Apart from self-attention, we incorporate information from the encoder representation of the current source, $X$, using decoder style cross-attention.
\item The retrieved target, $Y^i$, is encoded in a similar manner, attending the encoded representation of $X^i$ generated in the previous step.
\end{itemize}
The encoded representations for all targets, $\{Y^i,\, 1 \leq i \leq N\}$, are then concatenated along the time axis to form the Conditional Source Target Memory (CSTM).

\subsubsection{Gated Multi-Source Attention}
\label{subsubsec_gmsd}
We use gated multi-source attention to combine the context from the source encoder representations and the CSTM. This is similar to the gated attention employed by \cite{D18-1340}. We use a Transformer based decoder that attends to both, the encoder outputs and the CSTM, in every cross-attention layer. The rest of the decoder architecture remains unchanged.

Let the context vectors obtained by applying multi-head attention to the source and memory, with query $q_t$ be $c_t^s$ and $c_t^m$ respectively. Then the gated context vector, $c^t$, is given by,
\begin{equation}
    g_t = \sigma(W_{gs}c_t^s + W_{gm}c_t^m)
\end{equation}
\begin{equation}
    c_t = g_t*c_t^s + (1-g_t)*c_t^m
\end{equation}
where $g_t$ is the scalar gating variable at time-step t, and $W_{gs}$ and $W_{gm}$ are learned parameters. These steps are illustrated in Figure~\ref{fig:gmsd}.

\section{Experiments}
\label{sec_exp}
\subsection{Data and Evaluation}
We compare the performance of a standard Transformer Base model and our semi-parametric NMT approach on an English-French translation task. We create a new heterogeneous dataset, constructed from a combination of the WMT training set (36M pairs), the IWSLT bilingual corpus (237k pairs), JRC-Acquis (797k pairs)\footnote{From http://opus.nlpl.eu/JRC-Acquis.php} and OpenSubtitles (33M pairs)\footnote{From http://opus.nlpl.eu/OpenSubtitles.php}. For WMT, we use newstest 13 for validation and newstest 14 for test. For IWSLT, we use a combination of the test corpora from 2012-14 for validation and test 2015 for eval. For OpenSubtitles and JRC-Acquis, we create our own splits for validation and test, since no benchmark split is publicly available. After deduping, the JRC-Acquis test and validation set contain 6574 and 5121 sentence pairs respectively. The OpenSubtitles test and validation sets contain 3975 and 3488 pairs. For multi-domain training, the validation set is a concatenation of the four individual validation sets.

\begin{table*}[h]
\begin{center}
\begin{tabular}{l|l||l|l|l|l}
Model & Data         & newstest 14 & IWSLT 2015 & OpenSub & JRC-Acquis \\\hline \hline
TransformerBase & Multi Domain (MD) &       41.92        &     43.17     &        26.67        &      56.19   \\\hline
+ CSTM & MD + IDF Sentence  &        40.89 & 42.35        &	28.25&      65.38       \\\hline
+ CSTM & MD + IDF N-Gram    &          41.92 & \textbf{45.09} & 28.74     &      66.39        \\\hline
+ CSTM & MD + Dense N-Gram  &          \textbf{42.41} & \textbf{45.02}     & \textbf{29.06}  &     \textbf{66.92}    \\\hline       
\end{tabular}
\caption{Comparison of test translation quality (BLEU) with different retrieval strategies. Multi-domain is a concatenation of all 4 datasets. IDF Sentence, IDF-NGram and Dense N-Gram correspond to multi-domain datasets constructed with the different retrieval strategies.\label{tab:retrievaltest}}
\end{center}
\end{table*}

All datasets are tokenized with the Moses tokenizer \cite{koehn2007moses} and mixed without any sampling. We use a shared vocabulary Sentence-Piece Model \cite{kudo2018sentencepiece} for sub-word tokenization, with a vocabulary size of 32000 tokens. We train each model for 1M steps, and choose the best checkpoint from the last 5 checkpoints based on validation performance. BLEU scores are computed with tokenized true-cased output and references with \textit{multi-bleu.perl} from Moses.

\begin{table*}[h]
\begin{center}
\begin{tabular}{p{3cm}|p{12cm}}
source & `\color{orange}\underline{The top copy of the} \color{black}passenger waybill \color{green}\textit{shall be kept on the} \color{black}bus or coach throughout \color{purple}\textbf{the journey to which it refers .}\color{black}' \\\hline

neighbor source & `\color{orange}\underline{The top copy of the} \color{black}journey form \color{green}\textit{shall be kept on the }\color{black}vehicle during the whole of \color{purple}\textbf{the journey to which it refers .}\color{black}' \\\hline

baseline translation & `La  copie sup{\'e}rieure de la lettre de transport de voyageurs doit {\^e}tre conserv{\'e}e dans l' autobus ou l' autocar tout au long du voyage auquel elle se rapporte .\color{black}' \\\hline

neighbor target & `\color{orange}\underline{L' original de la feuille de route doit se trouver {\`a} bord} \color{black}du v{\'e}hicule \color{green}\textit{pendant toute la dur{\'e}e du voyage pour lequel elle a {\'e}t{\'e} {\'e}tablie .}\color{black}' \\\hline

translation & `\color{orange}\underline{L' original de la feuille de route doit se trouver {\`a} bord} \color{black}de l' autobus ou de l' autocar \color{green}\textit{pendant toute la dur{\'e}e du voyage pour lequel elle a {\'e}t{\'e} {\'e}tablie .}\color{black}' \\\hline

reference & `\color{orange}\underline{L' original de la feuille de route doit se trouver {\`a} bord} \color{black}de l' autobus ou de l' autocar \color{green}\textit{pendant toute la dur{\'e}e du voyage pour lequel elle a {\'e}t{\'e} {\'e}tablie .}\color{black}' \\\hline

\end{tabular}
\caption{A comparison of model outputs on a sample from the JRC-Acquis dataset. This model was trained using IDF based sentence level retrieval with Conditional Source Target Memory. The different colors and text formatting (underlined, italic, bold) represent different overlapping phrases within the model output, the retrieved target and the reference translation.\label{tab:jrc_eg}}
\end{center}
\end{table*}

For IDF based sentence retrieval, for each sentence in the training, dev and test corpus, we use $N=10$ neighbors per example during both, training and evaluation. For the N-Gram level retrieval strategies, we used $N=10$ neighbors during training, and neighbors corresponding to all n-grams during decoding. This was meant to limit memory requirements and enable the model to fit on P100s during training. We used n-gram width, $n=\{6, 10, 18\}$, for both IDF and dense vector based n-gram retrieval approaches. For scalability reasons, we restricted the retrieval set to the in-domain training corpus, i.e. neighbors for all train, dev and test sentences in the JRC-Acquis corpus were retrieved from the JRC-Acquis training split, and similarly for the other datasets.

\subsection{Hyper-parameters and Optimization}
For our baseline model we use the standard Transformer Base model \cite{DBLP:journals/corr/VaswaniSPUJGKP17}. For the semi-parametric model, all our hyper-parameters for attention (8 attention heads), model dimensions (512) and hidden dimensions (2048), including those used in the CSTM memory are equivalent to Transformer Base.

The Transformer baselines are trained on 16 GPUs, with the learning rate, warm-up schedule and batching scheme described in \cite{DBLP:journals/corr/VaswaniSPUJGKP17}. The semi-parametric models were trained on 32 GPUs with each replica split over 2 GPUs, one to train the translation model and the other for computing the CSTM. We used a conservative learning rate schedule (3, 40K) \cite{chen2018best} to train the semi-parametric models.

We apply a dropout rate\cite{srivastava2014dropout} of 0.1 to all inputs, residuals, attentions and ReLU connections in both models. We use Adam \cite{DBLP:journals/corr/KingmaB14} to train all models, and apply label smoothing with an uncertainty of 0.1 \cite{DBLP:journals/corr/SzegedyVISW15}. In addition to the transformer layers, layer normalization \cite{ba2016layer} was applied to the output of the CSTM. All models are implemented in Tensorflow-Lingvo \cite{lingvo}.

\begin{table*}[h]
\begin{center}
\begin{tabular}{p{3cm}|p{12cm}}
source & `Consciousness also is \color{orange}\textbf{what makes life worth living .}\color{black}'\\\hline

neighbor source & `So in the last 10 years and the hope for the future , we 've seen the beginnings of a science of positive psychology , a science of \color{orange}\textbf{what makes life worth living .}\color{black}'\\\hline

baseline translation & `La conscience est aussi ce \color{red}\textit{qui rend la vie valable .}\color{black}'\\\hline

neighbor target & `Donc , depuis 10 ans , et , esp{\'e}rons-le , {\`a} l' avenir nous assistons {\`a} l' {\'e}mergence d' une science de la psychologie positive : une science qui fait en sorte \color{orange}\textbf{que la vie} \color{black}vaille \color{orange}\textbf{la peine d' {\^e}tre v{\'e}cue .}\color{black}'\\\hline

translation & `La conscience est aussi ce qui fait \color{orange}\textbf{que la vie} \color{black}vaut \color{orange}\textbf{la peine d' {\^e}tre v{\'e}cue .}\color{black}'\\\hline

reference & `La conscience est aussi ce qui fait \color{orange}\textbf{que la vie} \color{black}vaut \color{orange}\textbf{la peine d' {\^e}tre v{\'e}cue .}\color{black}'\\\hline
\end{tabular}
\caption{A comparison of model outputs on a sample from IWSLT. This model was trained using IDF based n-gram retrieval with Conditional Source Target Memory. N-Gram level retrieval results in finding neighbors with high local overlap, even when the rest of the sentences are dissimilar.\label{tab:iwslt_eg}}
\end{center}
\end{table*}

\begin{table*}[h]
\begin{center}
\begin{tabular}{p{3cm}|p{12cm}}
source & `I was expecting to see \color{orange}\textbf{gnashing of teeth} \color{black}and a fight breaking out at the gate .\color{black}'\\\hline

neighbor source & `One could almost hear the collective \color{orange}\textbf{gnashing of teeth} \color{black}in the US , especially in the Congress .\color{black}'\\\hline

baseline translation & `J' esp{\'e}rais voir \color{red}\textit{des dents br{\^u}lantes }\color{black}et une bataille {\'e}clater {\`a} la porte .\color{black}'\\\hline

neighbor target & `On a presque entendre \color{orange}\textbf{les dents grincer} \color{black}aux {\'E}tats-Unis , surtout au Congr{\`e}s .\color{black}'\\\hline

translation & `Je m' attendais {\`a} voir \color{orange}\textbf{des grincements de dents} \color{black}et une bagarre {\'e}clater {\`a} la porte .\color{black}'\\\hline

reference & `Je m' attendais {\`a} voir \color{orange}\textbf{des grincements de dents} \color{black}et une bagarre {\'e}clater {\`a} la porte .\color{black}'\\\hline
\end{tabular}
\caption{A comparison of model outputs on a sample from WMT. This model was trained using IDF based n-gram retrieval with Conditional Source Target Memory. N-Gram level retrieval results in finding neighbors with high local overlap, even when the rest of the sentences are dissimilar.\label{tab:wmt_ng_eg}}
\end{center}
\end{table*}

\begin{table*}[h]
\begin{center}
\begin{tabular}{p{3cm}|p{12cm}}
source & `The artist \color{orange}\textbf{died} \color{black}last Sunday at the age of 71 .\color{black}'\\\hline

neighbor source & `A former minister George Thomson \color{orange}\textbf{passed away} \color{black}last week at the age of 87 .\color{black}'\\\hline

baseline translation & `L' artiste est \color{red}\textit{mort} \color{black}dimanche dernier {\`a} l' {\^a}ge de 71 ans .\color{black}'\\\hline

neighbor target & `George Thomson , ancien ministre , est \color{orange}\textbf{d{\'e}c{\'e}d{\'e}} \color{black}la semaine derni{\`e}re {\`a} l' {\^a}ge de 87 ans .\color{black}'\\\hline

translation & `L' artiste est \color{orange}\textbf{d{\'e}c{\'e}d{\'e}} \color{black}dimanche dernier {\`a} l' {\^a}ge de 71 ans .\color{black}'\\\hline

reference & `L' artiste est \color{orange}\textbf{d{\'e}c{\'e}d{\'e}} \color{black}dimanche dernier , {\`a} l' {\^a}ge de 71 ans . \color{black}'\\\hline
\end{tabular}
\caption{A comparison of model outputs on a sample from WMT. This model was trained using dense vector based n-gram retrieval with Conditional Source Target Memory. Dense vector based n-gram retrieval allows us to find semantically similar phrases, even when the lexical context is dissimilar.\label{tab:wmt_eg}}
\end{center}
\end{table*}

\section{Results}
\label{sec_res}
We compare the test performance of a multi-domain Transformer Base and our semi-parametric model using dense vector based n-gram retrieval and CSTM in Table~\ref{tab:retrievaltest}. Apart from significantly improving performance by more than 10 BLEU points on JRC-Acquis, 2-3 BLEU on OpenSubtitles and IWSLT, we notice a moderate gain of ~0.5 BLEU points on WMT 14. 
\begin{table*}[h]
\begin{center}
\begin{tabular}{l||l|l|l|l}
Model         & newstest 14 & IWSLT 2015 & OpenSub & JRC-Acquis \\\hline \hline
No Memory  &          41.92        &     43.17     &        26.67        &      56.19 \\\hline     
TM  &          41.64 &	44.32     &          27.38      &      64.25 \\\hline     
CTM    &         41.87       &       44.76     &        27.74        &       65.18      \\\hline 
CSTM  &          \textbf{42.41} & \textbf{45.02}     & \textbf{29.06}  &     \textbf{66.92}       \\\hline 
\end{tabular}
\caption{Comparison of test translation quality (BLEU) with different memory architectures. All models are trained on the Dense N-Gram Multi-Domain dataset. CSTM corresponds to the proposed Conditional Source Target Memory. CTM corresponds to Conditional Target Memory, where we ignore the retrieved sources while encoding the retrieved targets, and directly attend the encoding of the current source, $X$. TM corresponds to encoding the retrieved targets without any context.\label{tab:memory_ablation}}
\end{center}
\end{table*}

\subsection{Comparison of retrieval strategies}
We compare the performance of all 3 retrieval strategies in Table \ref{tab:retrievaltest}. The semi-parametric model with sentence level retrieval out-performs the seq2seq model by a huge margin on JRC-Acquis and OpenSubtitles. A sample from the JRC-Acquis dataset where the semi-parametric approach improves significantly over the neural approach is included in Table~\ref{tab:jrc_eg}. We notice that there is a lot of overlap between the source sentence and the retrieved source, resulting in the semi-parametric model copying large chunks from the retrieved target. However, its performance is noticeably worse on WMT and IWSLT. Based on a manual inspection of the retrieved candidates, we attribute these losses to retrieval failures. For broad domain datasets like WMT and IWSLT sentence retrieval fails to find good candidates.

Switching to n-gram level retrieval brings the WMT performance close to the seq2seq approach, and IWSLT performance to 2 BLEU points above the baseline model. Representative examples from IWSLT and WMT where n-gram retrieval improves over sentence level retrieval can be seen in Tables~\ref{tab:iwslt_eg} and \ref{tab:wmt_ng_eg}. Despite the majority of the retrieved neighbor having nothing in common with the source sentence, n-gram retrieval is able to find neighbors that contain local overlaps.

Using dense n-gram retrieval allows us to move beyond lexical overlap and retrieve semantically similar n-grams even when the actual tokens are different. As a result, dense n-gram retrieval improves performance over all our models on all 4 datasets. An illustrative example from WMT is included in Table~\ref{tab:wmt_eg}.

\subsection{Memory Ablation Experiments}
We report the performance of the various memory ablations in Table ~\ref{tab:memory_ablation}. We first remove the retrieved sources, $X^i$, from the CSTM, resulting in an architecture where the encoding of a retrieved target, $Y^i$, only incorporates information from the source $X$, represented by the row CTM in the table. This results in a clear drop in performance on all datasets. We ablate further by removing the attention to the original source $X$, resulting in a slightly smaller drop in performance (represented by TM). These experiments indicate that incorporating context from the sources significantly contributes to performance, by allowing the model to distinguish between relevant context and noise.

\begin{table*}[h]
\begin{center}
\begin{tabular}{l||l|l|l|l}
Adaptation Strategy & newstest 14 & IWSLT 2015 & OpenSub & JRC-Acquis \\\hline \hline
Base &       41.16          &    39.75    &       22.92         &      53.1\\ \hline
Fine-tuning  &     -            &       \textbf{42.87}     &         26.55       &       62.99       \\ \hline
Non-Parametric (NP) &     \textbf{41.57}            &       40.95     &         \textbf{27.09}       &       \textbf{64.93}       \\ \hline \hline
NP + Fine-tuning &     -            &       43.82     &         29.12       &       66.72       \\ \hline \hline
\end{tabular}
\caption{Comparison of test translation quality (BLEU) with different adaptation strategies. The base model (Transformer Base) is trained on the WMT dataset. Fine-tuning corresponds to fine-tuning based adaptation, where we initialize the domain-specific model from the WMT pre-trained Base model, and fine-tune it on the in-domain dataset for a few epochs. Non-parametric corresponds to our semi-parametric NMT model, adapted to in-domain data during inference by retrieving neighbors from the in-domain training corpus.\label{tab:fast_adapt}}
\end{center}
\end{table*}









\section{Non-Parametric Adaptation}
\label{sec_np}
Using a semi-parametric formulation for MT opens up the possibility of non-parametric adaptation. The biggest advantage of this approach is the possibility of training a single massively customizable model which can be adapted to any new dataset or document at inference time, by just updating the retrieval dataset.

We evaluate our model's performance on non-parametric adaptation and compare it against a fully fine-tuned model. In this setting, we train a baseline model and a dense n-gram based semi-parametric model on the WMT training corpus. We only retrieve and train on examples from the WMT corpus during training. We use the same hyper-parameters and training approaches used for the multi-domain experiments, as in Section \ref{sec_exp}.

The baseline model is then fine-tuned independently on JRC-Acquis, OpenSubtitles and IWSLT. The semi-parametric model is adapted non-parametrically to these three datasets, without any parameter updates. Adaptation is achieved via the retrieval mechanism - while evaluating, we retrieve similar examples from their respective training datasets. To quantify headroom, we also fine-tune our semi-parametric model on each of these datasets.

The results for non-parametric adaptation experiments are documented in Table ~\ref{tab:fast_adapt}. We notice that the non-parametric adaptation strategy significantly out-performs the base model on all 4 datasets. More importantly, the we find that our approach is capable of adapting to both, JRC-Acquis and OpenSubtitles, via just the retrieval apparatus, and out-performs the fully fine-tuned model indicating that non-parametric adaptation might be a reasonable approach when adapting to a lot of narrow domains or documents.

In-domain fine-tuning on top of non-parametric adaptation further improves by ~2 BLEU points on all datasets, increasing the gap even further with the seq2seq adapted models.

\section{Related Work}
Tools incorporating information from individual translation pairs, or translation memories \cite{lagoudaki2006translation,reinke2013state}, have been widely utilized by human translators in the industry. There have been a few efforts attempting to combine non-parametric methods with NMT \cite{gu2017search,zhang2018guiding,D18-1340}, but the key difference of our approach is the introduction of local, sub-sentence level similarity in the retrieval process, via n-gram level retrieval. Combined with our architectural improvements, motivated by the target encoder and gated attention from \cite{D18-1340} and the extended transformer model from \cite{D18-1049}, our semi-parametric NMT model is able to out-perform purely neural models in broad multi-domain settings.

Some works have proposed using phrase tables or the outputs of Phrase based MT within NMT  \cite{dahlmann2017neural,zhang2017improving,zhou2017neural}. While this reduces the noise present within the retrieved translation pairs, it requires training and maintaining a separate SMT system which might introduce errors of its own.

Another class of methods requires fine-tuning the entire NMT model to every instance at inference time, using retrieved examples \cite{farajian2017multi,wuebker2015hierarchical}, but these approaches require running expensive gradient descent steps before every translation.

Beyond NMT, there have been a few other attempts to incorporate non-parametric approaches into neural generative models \cite{guu2018generating,hayati2018retrieval,weston2018retrieve}. This strong trend towards combining neural generative models with non-parametric methods is an attempt to counter the weaknesses of neural networks, especially their failure to remember information from individual training instances and the diversity problem of seq2seq models \cite{vijayakumar2016diverse,jiang2018sequence}.

While our approach relies purely on retrieval from the training corpus, there has been quite a lot of work, especially on Question Answering, that attempts to find additional signals to perform the supervised task in the presence of external knowledge sources \cite{chen2017reading,wang2018r3}. Retrieving information from unsupervised corpora by utilizing multilingual representations \cite{guo2018effective} might be another interesting extension of this work.

\section{Conclusions and Future Work}
We make two major technical contributions in this work which enable us to improve the quality of semi-parametric NMT on broad domain datasets. First, we propose using n-gram retrieval, with standard Inverse Document Frequency similarity and with dense vector representations, that takes into account local sentence similarities that are critical to translation. As a result we are able to retrieve useful candidates even for broad domain tasks with little train-test overlap. Second, we propose a novel architecture to encode retrieved source-target pairs, allowing the model to distinguish useful information from noise by encoding the retrieved targets in context of the current translation task.

We demonstrate, for the first time, that semi-parametric methods can beat neural models by significant margins on multi-domain Machine Translation. By successfully training semi-parametric neural models on a broad domain dataset (WMT), we also open the door for non-parametric adaptation, showing huge improvements on new domains without any parameter updates. 

While we constrain this work to retrieved context, our architecture can be utilized to incorporate information from other sources of context, including documents, bilingual dictionaries etc. Using dense representations for retrieval also allows extending semi-parametric neural methods to other input modalities, including images and speech.

With this work, we hope to motivate further investigation into semi-parametric neural models for and beyond Neural Machine Translation.

\section*{Acknowledgments}

We would like to thank Naveen Arivazhagan, Macduff Hughes, Dmitry Lepikhin, Mia Chen, Yuan Cao, Ciprian Chelba, Zhifeng Chen, Melvin Johnson and other members of the Google Brain and Google Translate teams for their useful inputs and discussions. We would also like to thank the entire Lingvo development team for their foundational contributions to this project. \\

\bibliography{naaclhlt2019}
\bibliographystyle{acl_natbib}

\appendix

\end{document}